\documentclass[10pt,twocolumn,letterpaper]{article}

\usepackage{cvpr}
\usepackage{times}
\usepackage{epsfig}
\usepackage{graphicx}
\usepackage{amsmath}
\usepackage{amssymb}
\usepackage{lipsum}
\usepackage{amsfonts}
\usepackage{multirow}
\usepackage{subcaption}
\usepackage{rotating}
\usepackage{pgfplots}
\usepackage{eurosym}      
\usepackage{filecontents} 
\usepackage{tikz}
\usepackage{authblk}

\newcommand\crule[3][black]{\textcolor{#1}{\rule{#2}{#3}}}
\definecolor{road}{rgb}{1.0, 0.08, 0.58}
\definecolor{side-walk}{rgb}{0.29, 0.0, 0.51}
\definecolor{parking}{rgb}{0.8, 0.6, 0.8}
\definecolor{car}{rgb}{0.0, 0.4, 0.65}
\definecolor{bicyclist}{rgb}{0.63,0.17,0.53}
\definecolor{pole}{rgb}{0.82, 0.78, 0.50}
\definecolor{vegetation}{rgb}{0.13, 0.55, 0.13}
\definecolor{terrain}{rgb}{0.55,0.81,0.35}
\definecolor{trunk}{rgb}{0.64,0.28,0.0}
\definecolor{building}{rgb}{0.83,0.65,0.0}
\definecolor{other-structure}{rgb}{0.82,0.48,0.0}
\definecolor{other-object}{rgb}{0.17,0.87,0.87}

\usepackage[pagebackref=true,breaklinks=true,letterpaper=true,colorlinks,bookmarks=false]{hyperref}

\def\cvprPaperID{4498} 

\newcommand*{\method}{$\mathbf{(AF)^2}$-S3Net}
\newcommand*{\miou}{$\mathbf{mIoU}$}
\newcommand*{\firstranking}{$\mathbf{1^{st}}$ }

\ifcvprfinal\fi
\begin{document}

\title{\method: Attentive Feature Fusion with Adaptive Feature Selection for Sparse Semantic Segmentation Network}


\author[1]{Ran Cheng}
\author[1]{Ryan Razani}
\author[1]{Ehsan Taghavi}
\author[1]{Enxu Li}
\author[1]{Bingbing Liu}

\affil[1]{Noah Ark's Lab, Huawei, Markham, ON, Canada}

\maketitle


\begin{abstract}
Autonomous robotic systems and self driving cars rely on accurate perception of their surroundings as the safety of the passengers and pedestrians is the top priority. Semantic segmentation is one of the essential components of road scene perception that provides semantic information of the surrounding environment. Recently, several methods have been introduced for $3$D LiDAR semantic segmentation. While they can lead to improved performance, they are either afflicted by high computational complexity, therefore are inefficient, or they lack fine details of smaller instances. To alleviate these problems, we propose \method, an end-to-end encoder-decoder CNN network for $3$D LiDAR semantic segmentation. We present a novel multi-branch attentive feature fusion module in the encoder and a unique adaptive feature selection module with feature map re-weighting in the decoder. Our \method~ fuses the voxel-based learning and point-based learning methods into a unified framework to effectively process the large 3D scene. Our experimental results show that the proposed method outperforms the state-of-the-art approaches on the large-scale SemanticKITTI benchmark, ranking \firstranking on the competitive public leaderboard competition upon publication.

\end{abstract}
\vspace{-5px}

\section{Introduction}

Understanding of the surrounding environment has been one of the most fundamental tasks in autonomous robotic systems. With the challenges introduced with recent technologies such as self-driving cars, a detailed and accurate understanding of the road scene has become a main part of any outdoor autonomous robotic system in the past few years. To achieve an acceptable level of road scene understanding, many frameworks benefit from image semantic segmentation, where a specific class is predicted for every pixel in the input image, giving a clear perspective of the scene. 

\begin{figure}[htbp!]
    \centering
    \includegraphics[width=0.49\textwidth]{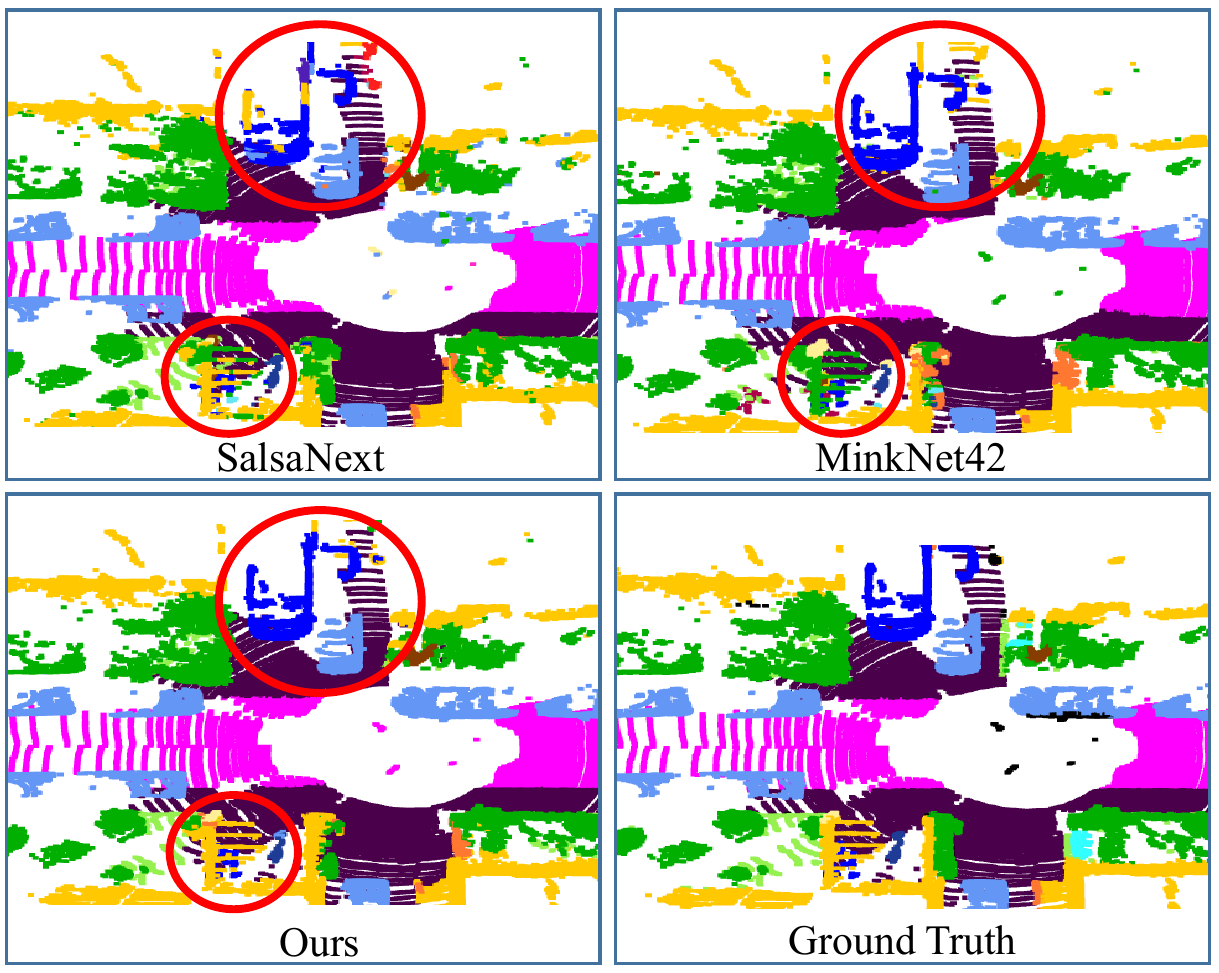}
    \caption{Comparison of our proposed method with SalsaNext \cite{cortinhal2020salsanext} and MinkNet42 \cite{choy20194d} on SemanticKITTI benchmark \cite{DBLP:conf/iccv/BehleyGMQBSG19}.
}
    \label{fig:kyber_intro}
    \vspace{-10px}
\end{figure}

Although image semantic segmentation is an important step in realizing self driving cars, the limitations of a vision sensor such as inability to record data in poor lighting conditions, variable sensor sensitivity, lack of depth information and limited field-of-view (FOV) makes it difficult for vision sensors to be the sole primary source for scene understanding and semantic segmentation. In contrast, Light Detection and Ranging (LiDAR) sensors can record accurate depth information regardless of the lighting conditions with high density and frame rate, making it a reliable source of information for critical tasks such as self driving.

LiDAR sensor generates point cloud by scanning the environment and calculating time-of-flight for the emitted laser beams. In doing so, LiDARs can collect valuable information, such as range (e.g., in Cartesian coordinates) and intensity (a measure of reflection from the surface of the objects). Recent advancement in LiDAR technology makes it possible to generate high quality, low noise and dense scans from desired environments, making the task of scene understanding a possibility using LiDARs. Although rich in information, LiDAR data often comes in an unstructured format and partially sparse at far ranges. These characteristics make the task of scene understating challenging using LiDAR as primary sensor. Nevertheless, research in scene understanding and in specific, semantic segmentation using LiDARs, has seen an increase in the past few years with the availability of datasets such as semanticKITTI \cite{DBLP:conf/iccv/BehleyGMQBSG19}.

The unstructured nature and partial sparsity of LiDAR data brings challenges to semantic segmentation. However, a great effort has been put by researchers to address these obstacles and many successful methods have been proposed in the literature (see Section \ref{sec:related_work}). From real-time methods which use projection techniques to benefit from the available $2$D computer vision techniques, to fully $3$D approaches which target higher accuracy, there exist a range of methods to build on. To better process LiDAR point cloud in $3$D and to overcome limitations such as non-uniform point densities and loss of granular information in voxelization step,  we propose \textbf{\method}, which is built upon Minkowski Engine \cite{choy20194d} to suit varying levels of sparsity in LiDAR point clouds, achieving state-of-the-art accuracy in semantic segmentation methods on SemanticKITTI \cite{DBLP:conf/iccv/BehleyGMQBSG19}. Fig.~\ref{fig:kyber_intro} demonstrates qualitative results of our approach compared to SalsaNext \cite{cortinhal2020salsanext} and MinkNet42 \cite{choy20194d}. We summarize our contributions as,

\begin{itemize}
  \itemsep0em
  \item An end-to-end encoder-decoder $3$D sparse CNN that achieves state-of-the-art accuracy in semanticKITTI benchmark \cite{DBLP:conf/iccv/BehleyGMQBSG19}; 
  \item A multi-branch attentive feature fusion module in the encoder to learn both global contexts and local details;
  \item An adaptive feature selection module with feature map re-weighting in the decoder to actively emphasize the contextual information from feature fusion module to improve the generalizability;
  \item A comprehensive analysis on semantic segmentation and classification performance of our model as opposed to existing methods on three benchmarks, semanticKITTI \cite{DBLP:conf/iccv/BehleyGMQBSG19}, nuScenes-lidarseg \cite{caesar2020nuscenes}, and ModelNet$40$ \cite{wu20153d} through ablation studies, qualitative and quantitative results.
\end{itemize}





\section{Related Work}
\label{sec:related_work}

\subsection{2D semantic Segmentation}

SqueezeSeg \cite{wu2018squeezeseg} is one of the first works on LiDAR semantic segmentation using range-image, where LiDAR point cloud projected on a $2$D plane using spherical transformation. SqueezeSeg \cite{wu2018squeezeseg} network is based on an encoder-decoder using Fully Connected Neural Network (FCNN) and a  Conditional Random Fields (CRF) as a Recurrent Neural Network (RNN) layer. In order to reduce number of the parameters in the network, SqueezeSeg incorporates ``fireModules'' from \cite{iandola2016squeezenet}. In a subsequent work, SqueezeSegV2 \cite{wu2019squeezesegv2} introduced Context Aggregation Module (CAM), a refined loss function and batch normalization to further improve the model. SqueezeSegV3 \cite{xu2020squeezesegv3} stands on the shoulder of \cite{wu2018squeezeseg, iandola2016squeezenet}, adopting a  Spatially-Adaptive Convolution (SAC) to use different filters in different locations in relation to the input image. Inspired by YOLOv3 \cite{redmon2018yolov3}, RangeNet++ \cite{milioto2019rangenet++} uses a DarkNet backbone to process a range-image. In addition to a novel CNN, RangeNet++ \cite{milioto2019rangenet++} proposes an efficient way of predicting labels for the full point cloud using a fast implementation of K-nearest neighbour (KNN).  

Benefiting from a new $2$D projection, PolarNet \cite{zhang2020polarnet} takes on a different approach using a polar Birds-Eye-View (BEV) instead of the standard $2$D grid-based BEV projections. Moreover, PolarNet encapsulates the information regarding each polar gird using PointNet, rather than using hand crafted features, resulting in a data-driven feature extraction, a nearest-neighbor-free method and a balanced  grid distribution. Finally, in a more successful attempt, SalsaNext \cite{cortinhal2020salsanext}, makes a series of improvements to the backbone introduced in SalsaNet \cite{salsanet2020} such as, a new global contextual block, an improved encoder-decoder and Lov\'asz-Softmax loss \cite{berman2018lovasz} to achieve state-of-the-art results in $2$D LiDAR semantic segmentation using range-image input.

\subsection{3D semantic Segmentation}
The category of large scale $3$D perception methods kicked off by early works such as \cite{shapenet2015,maturana2015voxnet,qi2016volumetric,wang2019voxsegnet,zhou2018voxelnet} in which a voxel representation was adopted to capitalize vanilla $3$D convolutions. In attempt to process unstructured point cloud directly, PointNet \cite{qi2017pointnet} proposed a Multi-Layer Perception (MLP)
to extract features from input points without any voxelization. PointNet++ \cite{qi2017pointnet++} which is an extension to the nominal work Pointnet \cite{qi2017pointnet}, introduced sampling at different scales to extract relevant features, both local and global. Although effective for smaller point clouds, Methods rely on Pointnet \cite{qi2017pointnet} and its variations are slow in processing large-scale data.

Down-sampling is at the core of the method proposed in RandLA-Net \cite{hu2020randla}. As down-sampling removes features randomly,  a local feature aggregation module is also introduced to progressively increase the receptive field for each 3D point. The two techniques used jointly to achieve both efficiency and accuracy in large-scale point cloud semantic segmentation. In a different approach, Cylinder$3$D \cite{zhou2020cylinder3d} uses cylindrical grids to partition the raw point cloud. To extract features, authors in \cite{zhou2020cylinder3d} introduced two new CNN blocks. An asymmetric residual block to ensure features related to cuboid objects are being preserved and Dimension-decomposition based Context Modeling in which multiple low-rank contexts are merged to model a high-ranked tensor suitable for $3$D point cloud data. 

Authors in KPConv \cite{thomas2019kpconv} introduced a new point convolution without any intermediate steps taken in processing point clouds. In essence, KPConv is a convolution operation which takes points in the neighborhood as input and processes them with spatially located weights. Furthermore, a deformable version of this convolution operator was also introduced that learns local shifts to make them adapt to point cloud geometry. Finally,  MinkowskiNet \cite{choy20194d} introduces a novel $4$D sparse convolution for spatio-temporal $3$D point cloud data along with an open-source library to support auto-differentiation for sparse tensors. Overall, where we consider the accuracy and efficiency, voxel-based methods such as MinkowskiNet \cite{choy20194d} stands above others, achieving state-of-the-art results within all sub-categories of $3$D semantic segmentation.

\subsection{Hybrid Methods}
Hybrid methods, where a mixture of voxel-based, projection-based and/or point-wise operations are used to process the point cloud, has been less investigated in the past, but with availability of more memory efficient designs, are becoming more successful in producing competitive results. For example, FusionNet \cite{zhang12356deep} uses a voxel-based MLP, called \textit{voxel-based mini-PointNet} which  directly aggregates features from all the points in the neighborhood voxels to the target voxel. This allows  FusionNet \cite{zhang12356deep} to search neighborhoods with low complexity, processing large scale point cloud with acceptable performance. In another approach, $3$D-MiniNet \cite{alonso2020MiniNet3D} proposes a learning-based projection module to extract local and global information from the $3$D data and then feeds it to a $2$D FCNN in order to generate semantic segmentation predictions. In a slightly different approach, MVLidarNet \cite{chen2020mvlidarnet} benefits form range-image LiDAR semantic segmentation to refine object instances in bird's-eye-view perspective, showcasing the applicability of LiDAR semantic segmentation in real-world applications. 

Finally, SPVNAS \cite{tang2020searching} builds upon the Minkowski Engine \cite{choy20194d} and designs a hybrid approach of using $4$D sparse convolution and point-wise operations to achieve state-of-the-art results in LiDAR semantic segmentation. To do this, authors in SPVNAS \cite{tang2020searching} use a neural architecture search (NAS) \cite{liu2018progressive} to efficiently design a NN, based on their novel Sparse Point-Voxel Convolution (SPVConv) operation.



\begin{figure*}[htbp!]
    \centering
    \includegraphics[width=\textwidth]{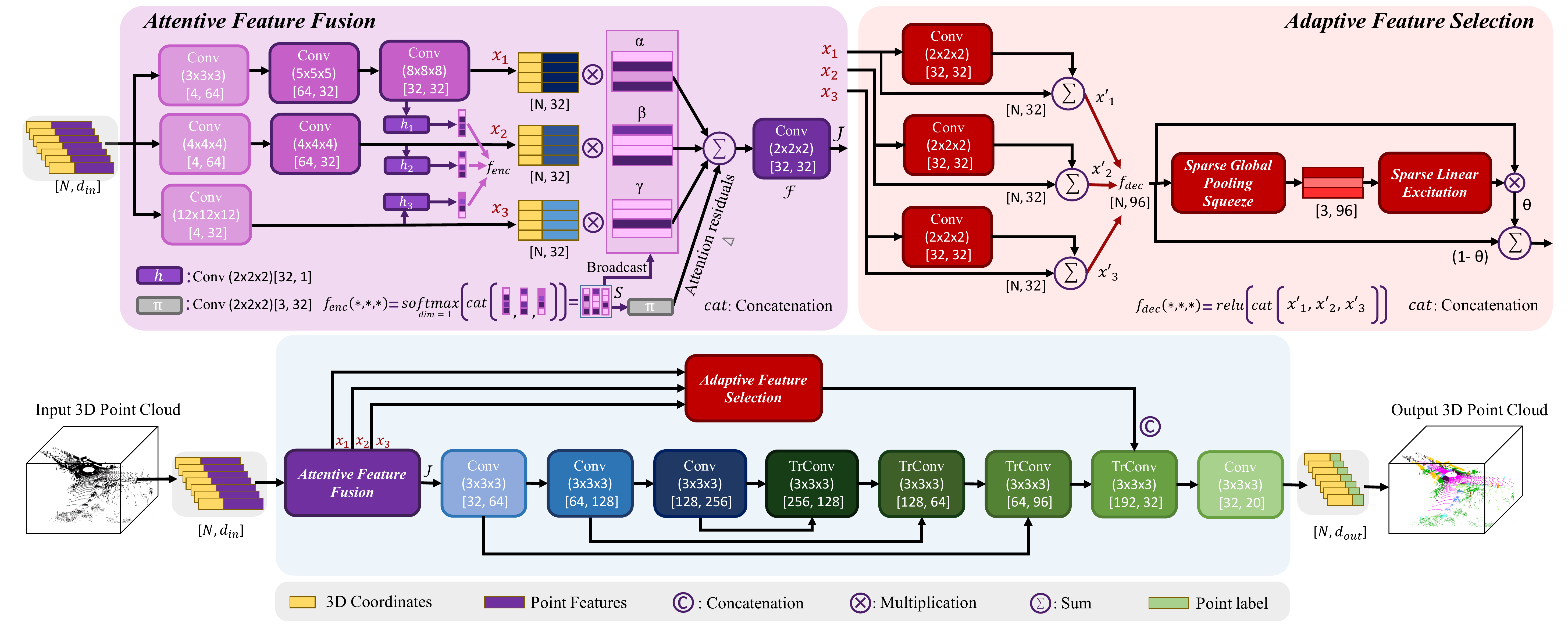}
    \caption{Overview of (AF)$^2$-S3Net. The top left block is Attentive Feature Fusion Module (AF2M) that aggregates Local and global context using a weighted combination of mutually exclusive learnable masks, $\alpha$, $\beta$, and $\gamma$. The top right block illustrates how Adaptive Feature Selection Module (AFSM) uses shared parameters to learn inter relationship between channels across multi-scale feature maps from AF2M. (best viewed on display)
 }
    \label{fig:S3ANet}
\end{figure*}

\section{Proposed Approach}
\label{sec:method}
The sparsity of outdoor-scene point clouds makes it difficult to extract spatial information compared to indoor-scene point clouds with fixed number of points or based on the dense image-based dataset. Therefore, it is difficult to leverage the indoor-scene or image-based segmentation methods to achieve good performance on a large-scale driving scene covering more than $100$m with non-uniform point densities. Majority of the LiDAR segmentation methods attempt to either transform 3D LiDAR point cloud into $2$D image using spherical projection (i.e., perspective, bird-eye-view) or directly process the raw point clouds. The former approach abandons valuable $3$D geometric structures and suffers from information loss due to projection process. The latter approach requires heavy computations and not feasible to be deployed in constrained systems with limited resources. Recently, sparse $3$D convolution became popular due to its success on outdoor LiDAR semantic segmentation task. However, out of a few methods proposed in \cite{choy20194d, tang2020searching}, no advanced feature extractors were proposed to enhance the results similar to computer vision and $2$D convolutions.

To overcome this, we propose (AF)$^2$-S3Net for LiDAR semantic segmentation in which a baseline model of MinkNet42 \cite{choy20194d} is transformed into an end-to-end encoder-decoder with attention blocks and achieves stat-of-the-art results. In this Section we first present the proposed network architecture along with its novel components, namely AF2M and AFSM. Then, the network optimization is introduced followed by the training details.

\subsection{Problem statement}

Lets consider a semantic segmentation task in which a LiDAR point cloud frame is given with a set of unordered points $(P, L) = (\{ p_{i},  l_{i} \})$ with $p_{i} \in \mathbb{R}^{d_{in}}$ and $i = 1, ... , N$, where $N$ denotes the number of points in an input point cloud scan. Each point $p_{i}$ contains $d_{in}$ input features, i.e., Cartesian coordinates $(x,y,z)$, intensity of returning laser beam $(i)$, colors $(R,G,B)$, etc.  Here, $l_{i} \in  \mathbb{R}$ represents the ground truth labels corresponding to each point $p_{i}$. However, in object classification task, a single class label $\mathcal{L}$ is assigned to an individual scene containing $P$ points.

Our goal is to learn a function $\mathfrak{F}_{cls}(., \Phi)$ parameterized by $\Phi$ that assigns a single class label $\mathcal{L}$ for all the points in the point cloud or in other words, $\mathfrak{F}_{seg}(., \Phi)$, that assigns a per point label $\hat{c}_{i}$ to each point $p_{i}$. To this end, we propose \method~ to minimize the difference between the predicted label(s), $\hat{\mathcal{L}}$ and $\hat{c}_{i}$, and the ground truth class label(s), $\mathcal{L}$ and $l_{i}$, for the tasks of classification and segmentation, respectively.

\subsection{Network architecture}

The block diagram of the proposed method, \method, is illustrated in Fig.~\ref{fig:S3ANet}. \method~ consists of a residual network based backbone and two novel modules, namely Attentive Feature Fusion module (AF2M) and Adaptive Feature Selection Module (AFSM). The model takes in a 3D LiDAR point cloud and transforms it into sparse tensors containing coordinates and features corresponding to each point. Then, the input sparse tensor is processed by \method~ which is built upon $3$D sparse convolution operations which suits sparse point clouds and effectively predicts a class label for each point given a LiDAR scan. 

A sparse tensor can be expressed as $P_s =[C, F]$, where $C \in \mathbb{R}^{N \times M}$ represents the input coordinate matrix with $M$ coordinates and $F \in \mathbb{R}^{N \times K}$ denotes its corresponding feature matrix with $K$ feature dimensions. 
In this work, we consider $3$D coordinates of points $(x,y,z)$, as our sparse tensor coordinate $C$, and per point normal features $(n_x,n_y,n_z)$ along with intensity of returning laser beam $(i)$ as our sparse tensor feature $F$. Exploiting  normal features helps the model to learn additional directional information, hence, the model performance can be improved by differentiating the fine details of the objects. The detailed description of the network architecture is provided below.

\begin{figure}[htbp!]
    \centering
    \includegraphics[width=0.45\textwidth]{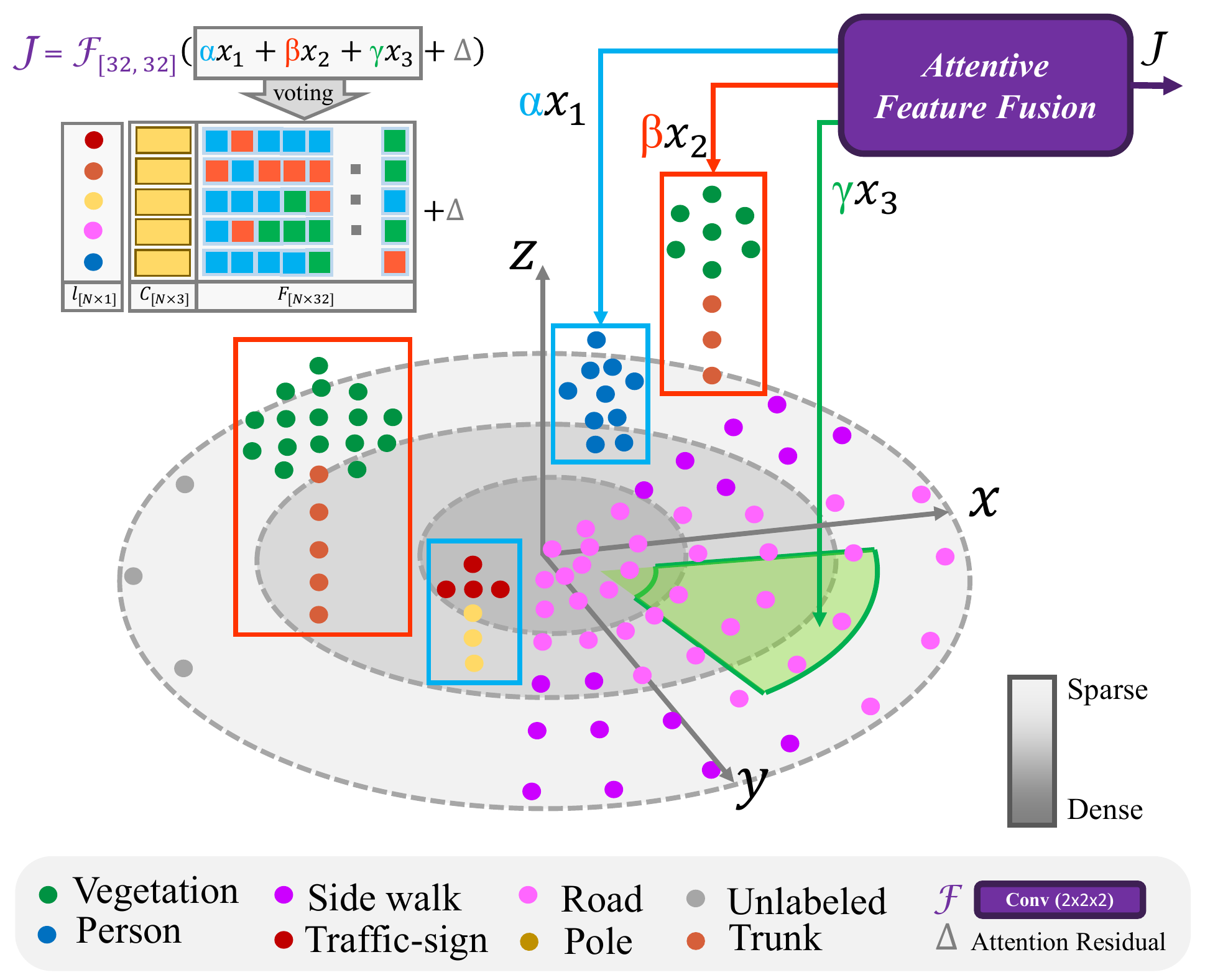}
    \caption{Illustration of Attentive Feature Fusion and spatial geometry of point cloud. The three labels $\alpha x_{1}$, $\beta x_{2}$, and $\gamma x_{3}$ represent the branches in AF2M encoder block. The first branch, $\alpha x_{1}$, learns to capture and emphasize the fine details for smaller instances such as person, pole and traffic-sign across the driving scenes with varying point densities. The shallower branches, $\beta x_{2}$ and $\gamma x_{3}$, learn different attention-maps that focus on global contexts embodied in larger instances such as vegetation, sidewalk and road surface. (best viewed on display)
}
    \label{fig:AFSM}
    \vspace{-10px}
\end{figure}

\textbf{Attentive Feature Fusion (AF2M)}: To better extract the global contexts, AF2M embodies a hybrid approach, covering small, medium and large kernel sizes, which focuses on point-based, medium-scale voxel-based and large-scale voxel-based features, respectively. The block diagram of AF2M is depicted in Fig.~\ref{fig:S3ANet} (top-left). 
Principally, the proposed AF2M fuses the features $\bar{x} = [x_1, x_2, x_3]$ at the corresponding branches using $g(\cdot)$ which is defined as,
\begin{equation}
     g(x_1, x_2, x_3) \triangleq \alpha x_1 + \beta x_2 + \gamma x_3 + \Delta
\end{equation}
where $\alpha$, $\beta$ and $\gamma$ are the corresponding coefficients that scale the feature columns for each point in the sparse tensor, and are  processed by function $f_{enc}(\cdot)$ as shown in Fig.~\ref{fig:S3ANet}. Moreover, the attention residuals, $\Delta$, is introduced to stabilize the attention layers $h_{i}(\cdot), \forall i \in \{1,2,3\}$, by adding the residual damping factor. This damping factor is the output of residual convolution layer $\pi$. Further, function $\pi$ can be formulated as
\begin{equation}
\pi \triangleq sigmoid(bn(conv(f_{enc}(\cdot)))) 
\end{equation} 
Finally, the output of AF2M is generated by $\mathcal{F}(g(\cdot))$, where $\mathcal{F}$ is used to align the sparse tensor scale space with the next convolution block. As illustrated in Fig.~\ref{fig:S3ANet} (top-left), for each $h_i$, $\forall i \in \{1,2,3\}$, the corresponding gradient of weight $w_{h_i}$ can be computed as:
\begin{equation}
\label{eq:weight_update_1}
    w_{h_i} = w_{h_i} - \frac{\partial J}{\partial g}\frac{\partial g}{\partial f_{enc}}\frac{\partial f_{enc}}{\partial h_i} - \frac{\partial J}{\partial g}\frac{\partial g}{\partial \pi}\frac{\partial \pi}{\partial h_i}
\end{equation}
where $J$ is the output of $\mathcal{F}$. Considering $g(\cdot)$ is a linear function of concatenated features $\bar{x}$ and $\Delta$, we can rewrite Eq. \ref{eq:weight_update_1} as follows:
\begin{equation}
    w_{h_i} = w_{h_i} - \frac{\partial J}{\partial g}\bar{x}\frac{\partial f_{enc}}{\partial h_i} - \frac{\partial J}{\partial g}\frac{\partial \pi}{\partial h_i}
\end{equation}
where $\frac{\partial f_{enc}}{\partial h_i} = \mathcal{S}_j(\mathcal{\delta}_{ij} - \mathcal{S}_{j})$ is the Jacobian of softmax function $S(\bar{x}): \mathbb{R}^{N} \rightarrow \mathbb{R}^{N}$ and maps $i$th input feature column to $j$th output feature column, and $\delta$ is Kronecker delta function where $\delta_{i=j} = 1$ and $\delta_{i\neq j} = 0$. As shown in Eq.~\ref{eq:S},

\begin{equation}
    \mathcal{S}_j(\mathcal{\delta}_{ij} - \mathcal{S}_{j}) = \begin{bmatrix} 
    -S^2_{1} & S_1(1-S_2) & \dots \\
    \vdots & \ddots & \\
    S_N(1-S_1) &  \dots      & -S^2_{N} 
    \end{bmatrix}
    \label{eq:S}
\end{equation}
when the softmax output $S$ is close to $0$, the term $\frac{\partial f_{enc}}{\partial h_i}$ approaches to zero which prompts no gradient, and when $S$ is close to $1$, the gradient is close to identity matrix. As a result, when $S\to1$, all values in $\alpha$ or $\beta$ or $\gamma$ get very high confidence and the update of $w_{h_i}$ becomes:
\begin{equation}
    w_{h_i} = w_{h_i} - \frac{\partial J}{\partial g}\frac{\partial \pi}{\partial h_i} + \frac{\partial J}{\partial g}\bar{x}I
\end{equation}
and in the case of $S\to0$, the update gradient will only depends on $\pi$. 
Fig.~\ref{fig:AFSM} further illustrates the capability of the proposed AF2M and highlights the effect of each branch visually. 

\textbf{Adaptive Feature Selection module (AFSM)}: The block diagram of AFSM is shown in Fig.~\ref{fig:S3ANet} (top-right). In AFSM decoder block, the feature maps from multiple branches in AF2M, $x_{1}$, $x_{2}$, and $x_{3}$, are further processed by residual convolution units. The resulted output, $x_{1}^{'}$, $x_{2}^{'}$, and $x_{3}^{'}$, are concatenated, shown as $f_{dec}$, and are passed into a shared squeeze re-weighting network \cite{hu2018squeeze} in which different feature maps are voted. 
This module acts like an adaptive dropout that intentionally filters out several feature maps that are not contributing to the final results. Instead of directly passing through the weighted feature maps as output, we employed a damping factor $\theta=0.35$, to regularize the weighting effect. It is worth noting that the skip connection connecting the attentive feature fusion module branches to the last decoder block, ensures that the error gradient propagates back to the encoder branches for better learning stability. 

\subsection{Network Optimization}

We leveraged a linear combination of geo-aware anisotrophic \cite{li2019depth},
Exponential-log loss \cite{wong20183d} and Lov\'asz loss \cite{berman2018lovasz} to optimize our network. In particular, geo-aware anisotrophic loss is beneficial to recover the fine details in a LiDAR scene. Moreover, Exponential-log loss \cite{wong20183d} loss is used to further improve the segmentation performance by focusing on both small and large structures given a highly unbalanced dataset.



The geo-aware anisotrophic loss can be computed by,

\begin{align}
    {L}_{geo}(y,\hat{y}) = -\frac{1}{N}\sum_{i,j,k}\sum_{c=1}^{C}\frac{M_{LGA}}{\psi}y_{ijk,c}log\hat{y}_{ijk, c}
\end{align}
where $y$ and $\hat{y}$ are the ground truth label and predicted label. Parameter $N$ is the local tensor neighborhood and in our experiment, we empirically set it as $5$ (a 10 voxels size cube). Parameter $C$ is the semantic classes, $M_{LGA} = \sum^{\Psi}_{\psi=1}(c_p \oplus c_{q_{\psi}})$, defined in \cite{li2019depth}. We normalized local geometric anisotropy within the sliding window $\Psi$ of the current voxel cell $p$ and its neighbor voxel grid $q_{\psi} \in \Psi$.

Therefore, the total loss used to train the proposed network is given by,
\begin{align}
    {L}_{tot}(y,\hat{y}) = w_{1} {L}_{exp}(y,\hat{y}) + w_{2} {L}_{geo}(y,\hat{y}) + w_{3} {L}_{lov}(y,\hat{y}) 
\end{align}
where $w_{1}$, $w_{2}$, and $w_{3}$ denote the weights of Exponential-log loss \cite{wong20183d}, geo-aware anisotrophic, and Lov\'asz loss, respectively. They are set as 1, 1.5 and 1.5 in our experiments.


\section{Experimental results}
\label{sec:results}

\begin{figure*}[htbp] 
    \centering
    \includegraphics[width=\textwidth]{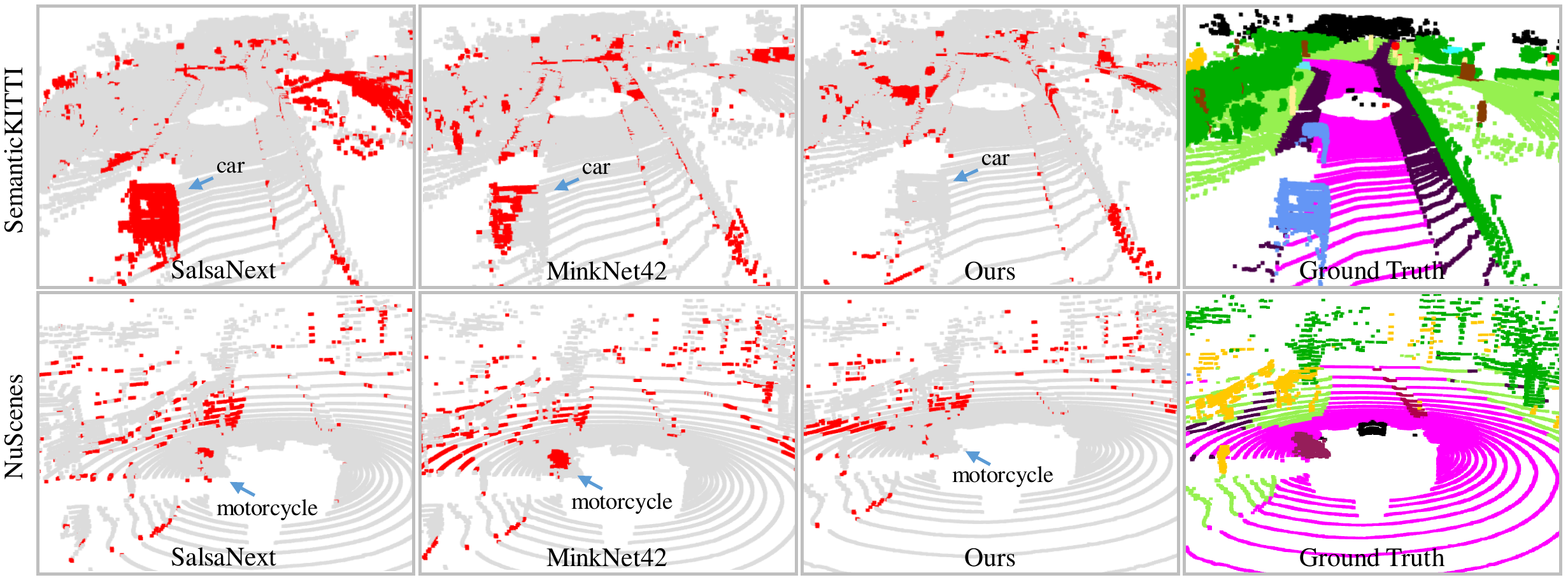}
    \caption{Compared to SalsaNext and MinkNet42, our method has a lower error (shown in red) recognizing region surface and smaller objects on nuScenes validation set, thanks to the proposed attention modules.}
    \label{fig:Skitti_nuScenes_error}
    \vspace{-10px}
\end{figure*}

We base our experimental results on three different dataset, namely, SemanticKITTI, nuScenes and ModelNet$40$ to show the applicability of the proposed methods in different scenes and domains. As for the SemanticKITTI and ModelNet$40$, \method~ is compared to the previous state-of-the-art, but due to a recently announce challenge for nuScenes-lidarseg dataset \cite{caesar2020nuscenes}, we provide our own evaluation results against the baseline model.  

To evaluate the performance of the proposed method and compare with others, we leverage mean Intersection over Union ($\mathbf{mIoU}$) as our evaluation metric. $\mathbf{mIoU}$ is the most popular metric for evaluating semantic point cloud segmentation and can be formalized as $mIoU = \frac{1}{n}\sum_{c=1}^{n}{\frac{TP_c}{TP_c+FP_c+FN_c}}$,
where $TP_c$ is the number of true positive points for class $c$, $FP_c$ is the number of false positives, and $FN_c$ is the number of false negatives. 


As for the training parameters, we trained our model with SGD optimizer with momentum of $0.9$ and learning rate of $0.001$, weight decay of $0.0005$ for $50$ epochs. The experiments are conducted using $8$ Nvidia V$100$ GPUs.

\subsection{Quantitative Evaluation}
In this Section, we provide quantitative evaluation of \method~on two outdoor large-scale public dataset: SemanticKITTI \cite{DBLP:conf/iccv/BehleyGMQBSG19} and nuScenes-lidarseg dataset \cite{caesar2020nuscenes} for semantic segmentation task and on ModelNet$40$ \cite{wu20153d} for classification task.

\begin{table*}[htbp!]
{\Huge
\centering
\resizebox{1.82\columnwidth}{!}{
\begin{tabular}{l|cccccccccccccccccccc}
\hline 
Method & \begin{sideways} Mean IoU \end{sideways} 
& \begin{sideways} Car \end{sideways} 
& \begin{sideways} Bicycle \end{sideways} 
& \begin{sideways} Motorcycle \end{sideways} 
& \begin{sideways} Truck \end{sideways} 
& \begin{sideways} Other-vehicle \end{sideways} 
& \begin{sideways} Person \end{sideways} 
& \begin{sideways} Bicyclist \end{sideways} 
& \begin{sideways} Motorcyclist \end{sideways} 
& \begin{sideways} Road \end{sideways} 
& \begin{sideways} Parking \end{sideways} 
& \begin{sideways} Sidewalk \end{sideways} 
& \begin{sideways} Other-ground \end{sideways} 
& \begin{sideways} Building \end{sideways} 
& \begin{sideways} Fence \end{sideways} 
& \begin{sideways} Vegetation \end{sideways} 
& \begin{sideways} Trunk \end{sideways} 
& \begin{sideways} Terrain \end{sideways} 
& \begin{sideways} Pole \end{sideways} 
& \begin{sideways} Traffic-sign \end{sideways}\\ 
\hline

S-BKI \cite{gan2020bayesian} 
& $51.3$ & $83.8$ & $30.6$ & $43.0$ & $26.0$ & $19.6$ & $8.5$ & $3.4$ & $0.0$ & $92.6$ & $65.3$ & $77.4$ & $30.1$ & $89.7$ & $63.7$ & $83.4$ & $64.3$ & $67.4$ & $58.6$ & $67.1$  \\
RangeNet++ \cite{milioto2019rangenet++} 
& $52.2$ & $91.4$ & $25.7$ & $34.4$ & $25.7$ & $23.0$ & $38.3$ & $38.8$ & $4.8$ & $91.8$ & $65.0$ & $75.2$ & $27.8$ & $87.4$ & $58.6$ & $80.5$ & $55.1$ & $64.6$ & $47.9$ & $55.9$  \\
LatticeNet  \cite{rosu2019latticenet} 
& $52.9$ & $92.9$ & $16.6$ & $22.2$ & $26.6$ & $21.4$ & $35.6$ & $43.0$ & $46.0$ & $90.0$ & $59.4$ & $74.1$ & $22.0$ & $88.2$ & $58.8$ & $81.7$ & $63.6$ & $63.1$ & $51.9$ & $48.4$ \\
RandLA-Net \cite{hu2020randla} 
& $53.9$ & $94.2$ & $26.0$ & $25.8$ & $40.1$ & $38.9$ & $49.2$ & $48.2$ & $7.2$ & $90.7$ & $60.3$ & $73.7$ & $20.4$ & $86.9$ & $56.3$ & $81.4$ & $61.3$ & $66.8$ & $49.2$ & $47.7$  \\
PolarNet \cite{zhang2020polarnet} 
& $54.3$ & $93.8$ & $40.3$ & $30.1$ & $22.9$ & $28.5$ & $43.2$ & $40.2$ & $5.6$ & $90.8$ & $61.7$ & $74.4$ & $21.7$ & $90.0$ & $61.3$ & $84.0$ & $65.5$ & $67.8$ & $51.8$ & $57.5$  \\
MinkNet42 \cite{choy20194d} 
& $54.3$ &  $94.3$ & $23.1$ & $26.2$ & $26.1$ & $36.7$ & $43.1$ & $36.4$ & $7.9$ & $91.1$ & $63.8$ & $69.7$ & $29.3$ & $\textbf{92.7}$ & $57.1$ & $83.7$ & $68.4$ & $64.7$ & $57.3$ & $60.1$ \\ 
3D-MiniNet \cite{alonso2020MiniNet3D} 
& $55.8$ & $90.5$ & $42.3$ & $42.1$ & $28.5$ & $29.4$ & $47.8$ & $44.1$ & $14.5$ & $91.6$ & $64.2$ & $74.5$ & $25.4$ & $89.4$ & $60.8$ & $82.8$ & $60.8$ & $66.7$ & $48.0$ & $56.6$ \\
SqueezeSegV3 \cite{xu2020squeezesegv3} 
& $55.9$ & $92.5$ & $38.7$ & $36.5$ & $29.6$ & $33.0$ & $45.6$ & $46.2$ & $20.1$ & $91.7$ & $63.4$ & $74.8$ & $26.4$ & $89.0$ & $59.4$ & $82.0$ & $58.7$ & $65.4$ & $49.6$ & $58.9$ \\
Kpconv \cite{thomas2019kpconv} 
& $58.8$ & $96.0$ & $30.2$ & $42.5$ & $33.4$ & $44.3$ & $61.5$ & $61.6$ & $11.8$ & $88.8$ & $61.3$ & $72.7$ & $31.6$ & $90.5$ & $64.2$ & $84.8$ & $69.2$ & $69.1$ & $56.4$ & $47.4$ \\ 
SalsaNext \cite{cortinhal2020salsanext}
& $59.5$ & $91.9$ & $48.3$ & $38.6$ & $38.9$ & $31.9$ & $60.2$ & $59.0$ & $19.4$ & $91.7$ & $63.7$ & $75.8$ & $29.1$ & $90.2$ & $64.2$ & $81.8$ & $63.6$ & $66.5$ & $54.3$ & $62.1$    \\
FusionNet   \cite{zhang12356deep}
& $61.3$  & $95.3$ & $47.5$ & $37.7$ & $41.8$ & $34.5$ & $59.5$ & $56.8$ & $11.9$ & $91.8$ & $68.8$ & $77.1$ & $30.8$ & $92.5$ & $\textbf{69.4}$ & $84.5$ & $69.8$ & $68.5$ & $60.4$ & $66.5$\\
KPRNet    \cite{kochanov2020kprnet}
& $63.1$ & $95.5$ & $54.1$ & $47.9$ & $23.6$ & $42.6$ & $65.9$ & $65.0$ & $16.5$ & $\textbf{93.2}$ & $\textbf{73.9}$ & $\textbf{80.6}$ & $30.2$ & $91.7$ & $68.4$ & $85.7$ & $69.8$ & $\textbf{71.2}$ & $58.7$ & $64.1$\\
SPVNAS    \cite{tang2020searching}
& $67.0$ & $\textbf{97.2}$ & $50.6$ & $50.4$ & $\textbf{56.6}$ & $\textbf{58.0}$ & $67.4$ & $67.1$ & $50.3$ & $90.2$ & $67.6$ & $75.4$ & $21.8$ & $91.6$ & $66.9$ & $\textbf{86.1}$ & $\textbf{73.4}$ & $71.0$ & $\textbf{64.3}$ & $67.3$\\
\hline
 \textbf{\method} [Ours] 
& $\textbf{69.7}$ & $94.5$ & $\textbf{65.4}$ & $\textbf{86.8}$ & $39.2$ & $41.1$ & $\textbf{80.7}$ & $\textbf{80.4}$ & $\textbf{74.3}$ & $91.3$ & $68.8$ & $72.5$ & $\textbf{53.5}$ & $87.9$ & $63.2$ & $70.2$ & $68.5$ & $53.7$ & $61.5$ & $\textbf{71.0}$
\\
\hline
\end{tabular}
}
\caption[S3Net]{Segmentation IoU (\%) results on the SemanticKITTI \cite{DBLP:conf/iccv/BehleyGMQBSG19} test dataset.}
\label{bigtable}}
\end{table*} 
\noindent \textbf{SemanticKITTI dataset:} we conduct our experiments on SemanticKITTI \cite{DBLP:conf/iccv/BehleyGMQBSG19} dataset, the largest dataset for autonomous vehicle LiDAR segmentation. This dataset is based on the KITTI dataset introduced in \cite{geiger2012cvpr}, containing $41000$ total frames which captured in $21$ sequences. We list our experiments with all other published works in Table \ref{bigtable}. As shown in Table \ref{bigtable}, our method achieves state-of-the-art performance in SemanticKITTI test set in terms of mean IoU. With our proposed method, \method, we see a $2.7\%$ improvement from the second best method \cite{tang2020searching} and $15.4\%$ improvement from baseline model (MinkNet42 \cite{choy20194d}). Our method dominates greatly in classifying small objects such as bicycle, person and motorcycle, making it a reliable solution to understating complex scenes. It is worth noting that \method~only uses the voxelized data as input, whereas the competing methods like SPVNAS \cite{tang2020searching} use both voxelized data and point-wise features.
\vspace{-3px}
\begin{table*}[htb!]
{\large
\centering
\resizebox{1.82\columnwidth}{!}{
\begin{tabular}{l|cccccccccccccccccc}
\hline 
Method 
& \begin{sideways} FW mIoU \end{sideways} 
& \begin{sideways} \textbf{Mean IoU} \end{sideways} 
& \begin{sideways} Barrier \end{sideways} 
& \begin{sideways} Bicycle \end{sideways} 
& \begin{sideways} Bus \end{sideways} 
& \begin{sideways} Car \end{sideways} 
& \begin{sideways}\begin{tabular}[l]{@{}l@{}}Construction\\ vehicle\end{tabular} \end{sideways} 
& \begin{sideways} Motorcycle \end{sideways} 
& \begin{sideways} Pedestrian \end{sideways} 
& \begin{sideways} Traffic cone \end{sideways} 
& \begin{sideways} Trailer \end{sideways} 
& \begin{sideways} Truck \end{sideways} 
& \begin{sideways} \begin{tabular}[l]{@{}l@{}}Driveable\\ surface\end{tabular} \end{sideways} 
& \begin{sideways} \begin{tabular}[l]{@{}l@{}}Other\\ flat ground\end{tabular} \end{sideways} 
& \begin{sideways} Sidewalk \end{sideways} 
& \begin{sideways} Terrain \end{sideways} 
& \begin{sideways} Manmade \end{sideways} 
& \begin{sideways} Vegetation \end{sideways} \\
\hline

SalsaNext \cite{cortinhal2020salsanext}
& $82.8 $ & $58.8 $ & $56.6 $ & $4.7 $ & $77.1 $ & $\textbf{81.0} $ & $18.4 $ & $47.5 $ & $52.8 $ & $43.5 $ & $38.3 $ & $65.7 $ & $94.2 $ & $60.0 $ & $\textbf{68.9} $ & $\textbf{70.3 }$ & $81.2 $ & $80.5$   \rule{0pt}{2ex} \\

MinkNet42 \cite{choy20194d} 
& $82.7 $ & $60.8 $ & $\textbf{63.1 }$ & $8.3 $ & $77.4 $ & $77.1 $ & $23.0 $ & $55.1 $ & $55.6 $ & $\textbf{50.0} $ & $\textbf{42.5} $ & $62.2 $ & $94.0 $ & $67.2 $ & $64.1 $ & $68.6 $ & $\textbf{83.7 }$ & $80.8 $ \\

\hline
 \textbf{\method~} [Ours] 
& $\textbf{83.0}$ & $\textbf{62.2 }$ & $60.3 $ & $\textbf{12.6} $ & $\textbf{82.3 }$ & $80.0 $ & $\textbf{20.1} $ & $\textbf{62.0} $ & $\textbf{59.0} $ & $49.0 $ & $42.2 $ & $\textbf{67.4} $ & $\textbf{94.2} $ & $\textbf{68.0 }$ & $64.1 $ & $68.6 $ & $82.9 $ & $\textbf{82.4}$   \\

\hline
\end{tabular}
}
\caption[Kybernet]{Segmentation IoU (\%) results on the nuScenes-lidarseg \cite{caesar2020nuscenes} validation dataset. Frequency-Weighted IoU denotes that each IoU is weighted by the point-level frequency of its class.}
\label{nuScenes}}
\end{table*} 

\noindent \textbf{Nuscenes dataset:} to prove the generalizability of our proposed method, we trained our network with nuScenes-lidarseg dataset \cite{caesar2020nuscenes}, one of the recently available large-scale datasets that provides point level labels of LiDAR point clouds. It consists of $1000$ driving scenes from various locations in Boston and Singapore, providing a rich set of labeled data to advance self driving car technology. Among these $1000$ scenes, $850$ of them is reserved for training and validation, and the remaining $150$ scenes for testing. The labels are, to some extent, similar to the semanticKITTI dataset \cite{DBLP:conf/iccv/BehleyGMQBSG19}, making it a new challenge to propose methods that can handle both datasets well, given the different sensor setups and environment they record the dataset. In Table \ref{nuScenes}, we compared our proposed method with MinkNet42 \cite{choy20194d} baseline and the projection based method SalsaNext \cite{cortinhal2020salsanext}. Results in Table~\ref{nuScenes} shows that our proposed method can handle the small objects in nuScenes dataset and indicates a large margin improvement from the competing methods. Considering the large difference between the two public datasets, we can prove that our work can generalize well.


\noindent \textbf{ModelNet$40$:} to expand and evaluate the capabilities of the proposed method in different applications, ModelNet$40$, a 3D object classification dataset \cite{wu20153d} is adopted for evaluation. ModelNet$40$ contains $12,311$ meshed CAD models from $40$ different object categories. From all the samples, $9,843$ models are used for training and $2,468$ models for testing. To evaluate our method against existing stat-of-the-art, we compare \method~ with techniques in which a single input (e.g., single view, sampled point cloud, voxel) has been used to train and evaluate the models. To make \method~ compatible for the task of classification, the decoder part of the network is removed and the output of the encoder is directly reshaped to the number of the classes in ModelNet$40$ dataset. Moreover, the model is trained only using cross-entropy loss. Table~\ref{tab_class_cpnet_results} presents the overall classification accuracy results for our proposed method and previous state-of-the-art. With the introduction of AF2M in our network, we achieved similar performance to the point-based methods which leverage fine-grain local features.  



\begin{table*}
\small
\centering
{
\begin{tabular}{l|c|c|c}
\hline
Method & Input & Main operator & Overall Accuracy (\%)   \\ \hline 
Vox-Net \cite{maturana2015voxnet} & voxels & 3D Operation & 83.00 \\ 
Mink-ResNet50 \cite{choy20194d} & voxels & Sparse 3D Operation & 85.30 \\ 
Pointnet \cite{qi2017pointnet} & point cloud & Point-wise MLP & 89.20 \\  
Pointnet++ \cite{qi2017pointnet++} & point cloud & Local feature & 90.70 \\  
DGCNN\cite{zhang2018end} (1 vote) & point cloud & Local feature & 91.84 \\
GGM-Net \cite{li2020ggm} & point cloud & Local feature & 92.60 \\   
RS-CNN \cite{liu2019relation} & point cloud & Local feature & \textbf{93.60} \\ 
\hline
Ours (AF2M) & voxels & Sparse 3D Operation & 93.16 \\ 
\hline
\end{tabular}
}
\caption[CPNET]{Classification accuracy results on ModelNet$40$ dataset \cite{wu20153d}, for input size $1024\times3$.}
\label{tab_class_cpnet_results}
\end{table*}

\subsection{Qualitative Evaluation}

In this section, we visualize the attention maps in AF2M by projecting the scaled feature maps back to original point cloud. Moreover, to better present the the improvements that has been made against the baseline model MinkNet$42$ \cite{choy20194d} and SalsaNext \cite{cortinhal2020salsanext}, we provide the error maps which highlights the superior performance of our method.

\begin{figure}[t]
    \centering
    \includegraphics[width=0.49\textwidth]{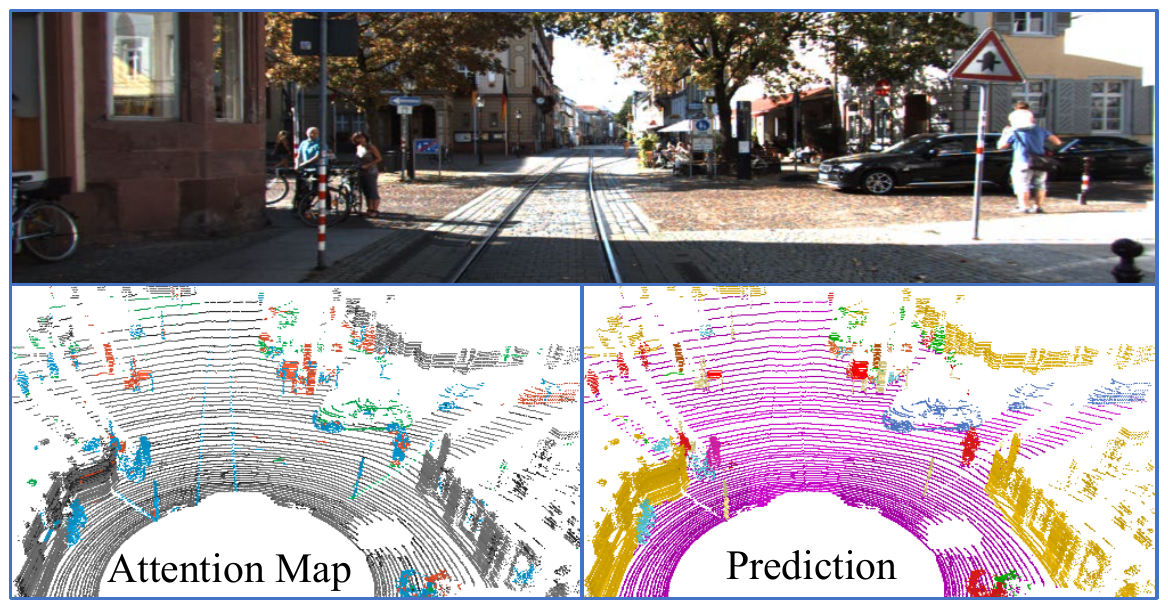}
    \caption{Reference image (top), Prediction (bottom-right), attention map (bottom-left) on SemanticKITTI test set.
    Color codes are: \crule[road]{0.2cm}{0.2cm} road \textbar
	\crule[side-walk]{0.2cm}{0.2cm} side-walk \textbar 
	\crule[parking]{0.2cm}{0.2cm} parking \textbar
	\crule[car]{0.2cm}{0.2cm} car \textbar 
	\crule[bicyclist]{0.2cm}{0.2cm} bicyclist \textbar
	\crule[pole]{0.2cm}{0.2cm} pole \textbar
	\crule[vegetation]{0.2cm}{0.2cm} vegetation \textbar
	\crule[terrain]{0.2cm}{0.2cm} terrain \textbar
	\crule[trunk]{0.2cm}{0.2cm} trunk \textbar
	\crule[building]{0.2cm}{0.2cm} building \textbar
	\crule[other-structure]{0.2cm}{0.2cm} other-structure \textbar
	\crule[other-object]{0.2cm}{0.2cm} other-object.}

    \label{fig:AF2M_at_map}
    \vspace{-10px}
\end{figure}

As shown in Fig.~\ref{fig:AF2M_at_map}, our method is capable of capturing fine details in a scene. To demonstrate this, we train \method~ on SemanticKITTI as explained above and visualize a test frame. In Fig.~\ref{fig:AF2M_at_map} we highlight the points with top 5\% feature norm from each scaled feature maps of $\alpha x_1$, $\beta x_2$ and $ \gamma x_3$ with cyan, orange and green colors, respectively. It can be observed that our model learns to put its attention on small instances (i.e., person, pole, bicycle, etc.) as well as larger instances (i.e., car, region boundaries, etc.).
Fig.~\ref{fig:Skitti_nuScenes_error} shows some qualitative results on SemanticKITTI (top) and nuScenes (bottom) benchmark. It can be observed that the proposed method surpasses the baseline (MinkNet$42$ \cite{choy20194d}) and range-based SalsaNext \cite{cortinhal2020salsanext} by a large margin, which failed to capture fine details such as cars and vegetation.

\subsection{Ablation Studies}
To show the effectiveness of the proposed attention mechanisms, namely, AF2M and AFSM introduced in Section \ref{sec:method}, along with other design choices such as loss functions, this section is dedicated to a thorough ablation study starting from our baseline model introduced in \cite{choy20194d}. The baseline is MinkNet$42$ which is a semantic segmentation residual NN model for $3$D sparse data. To start off with a well trained baseline, we use Exponential Logarithmic Loss \cite{wong20183d} to train the model which results in $59.8\%$ \miou~accuracy for the validation set on semanticKITTI.

Next, we add our proposed AF2M to the baseline model to help the model extract richer features from the raw data. This addition of AF2M improves the \miou~to $65.1\%$, an increase of $5.3\%$. In our second study and to show the effectiveness of the AFSM only, we first reduce the AF2M block to only output $\{x_1,x_2,x_3\}$ (see Fig.~\ref{fig:S3ANet} for reference), and then add the AFSM to the model. Adding AFSM shows an increase of $3.5\%$ in \miou~from the baseline. In the last step of improving the NN model, we combine AF2M and AFSM together as shown in Fig.~\ref{fig:S3ANet}, which result in \miou~of $68.6\%$ and an increase of $8.8\%$ from the baseline model. 

Finally, in our last two experiments, we study the effect of our loss function by adding Lov\'asz loss and the combination of Lov\'asz and geo-aware anisotrophic loss, resulting in \miou~of $70.2\% $ and $74.2\%$, respectively. The ablation studies presented, shows a series of adequate steps in the design of \method, proving the steps taken in the design of the proposed model are effective and can be used separately in other NN models to improve the accuracy.

\begin{table}[h]
\begin{center}
\scalebox{0.78}
{
\begin{tabular}{ c|cccc c  }
\hline 
\multicolumn{1}{c|}{\textbf{Architecture}} &
\multicolumn{1}{c}{\textbf{\begin{turn}{45} AF2M \end{turn} }} &
\multicolumn{1}{c}{\textbf{\begin{turn}{45} AFSM \end{turn} }} &
\multicolumn{1}{c}{\textbf{\begin{turn}{45} Lov\'asz \end{turn} }} &
\multicolumn{1}{c}{\textbf{\begin{turn}{45} Lov\'asz+Geo \end{turn} }} &
\multicolumn{1}{|c}{\textbf{mIoU}} \\

 \hline \hline 
\multirow{1}{*}{Baseline}
&     &   &       &   \multicolumn{1}{c|}{ } & 59.8 \rule{0pt}{3ex}\\\hline

\multirow{6}{*}{Proposed}
& \checkmark &   &           & \multicolumn{1}{c|}{ }  & 65.1  \rule{0pt}{3ex}\\ 
&  & \checkmark  &           & \multicolumn{1}{c|}{ }  & 63.3  \rule{0pt}{3ex}\\ 
& \checkmark & \checkmark &         &  \multicolumn{1}{c|}{ }  & 68.6  \rule{0pt}{3ex}\\ 


& \checkmark & \checkmark &\checkmark   &       \multicolumn{1}{c|}{ }  & 70.2  \rule{0pt}{3ex}\\ 

& \checkmark & \checkmark &\checkmark      &    \multicolumn{1}{c|}{\checkmark}  & 74.2  \rule{0pt}{3ex}\\ \hline

\end{tabular}}
\end{center}
\vspace{-10px}
\caption{Ablation study of the proposed method vs baseline evaluated on SemanticKITTI \cite{DBLP:conf/iccv/BehleyGMQBSG19} validation dataset (seq 08). }
\label{tab:Ablation}
\vspace{-10px}
\end{table}

\subsection{Distance-based Evaluation}
In this section, we investigate how segmentation is affected by distance of the points to the ego-vehicle. In order to show the improvements, we follow our ablation study and compare \method~ and the baseline (MinkNet42) on the SemanticKITTI validation set (seq $8$).  Fig.~\ref{range} illustrates the \miou~ of \method~ as opposed to the baseline and SalsaNext w.r.t. the distance to the ego-vehicle's LiDAR sensors. The results of all the methods get worse by increasing the distance due to the fact that point clouds generated by LiDAR are relatively sparse, especially at large distances. However, the proposed method can produce better results at all distances, making it an effective method to be deployed on autonomous systems. It is worth noting that, while the baseline methods attempt to alleviate the sparsity problem of point clouds by using sparse convolutions in a residual style network, it lacks the necessary encapsulation of features proposed in Section \ref{sec:method} to robustly predict the semantics.

\begin{center}
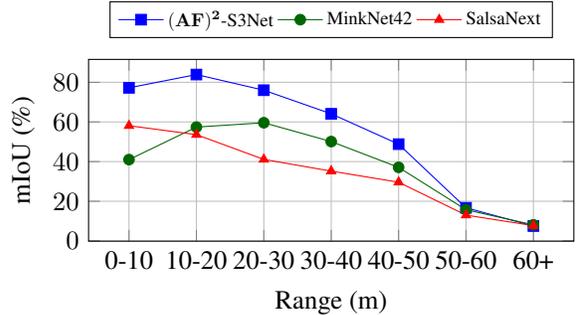

\begin{tikzpicture}
\begin{axis}[
    width=0.46\textwidth,
    height=4.0cm,
    grid=major,
    legend style={nodes={scale=0.7}, at={(0.5, 1.32)}, anchor=north, legend columns=-1},
	xlabel=Range (m),
	ylabel = mIoU (\%),
	ylabel near ticks,
    symbolic x coords={0-10,10-20,20-30,30-40,40-50,50-60,60+},
    xtick=data
]
\addplot[color=blue,mark=square*] table[x=interval,y=ioua] {interval-iou.csv};
\addplot[color=black!60!green,mark=*] table[x=interval,y=iouc] {interval-iou.csv};
\addplot[color=red,mark=triangle*] table[x=interval,y=ioub] {interval-iou.csv};

\legend{\method, MinkNet42, SalsaNext}
\end{axis} 
\end{tikzpicture}
\vspace{-10px}
\captionof{figure}{mIoU vs Distance for \method~ vs. baseline.} 
\label{range}
\vspace{-5px}
\end{center}

\section{conclusion}
\label{sec:conclusion}
In this paper, we presented an end-to-end CNN model to address the problem of semantic segmentation and classification of $3$D LiDAR point cloud. We proposed \method , a $3$D sparse convolution based network with two novel attention blocks called Attentive Feature Fusion Module (AF2M) and Adaptive Feature Selection Module (AFSM), to effectively learn local and global contexts and emphasize the fine detailed information in a given LiDAR point cloud.
Extensive experiments on several benchmarks, SemanticKITTI, nuScenes-lidarseg, and ModelNet40 demonstrated the ability to capture the local details and the state-of-the-art performance of our proposed model. Future work will include the extension of our method to end-to-end $3$D instance segmentation and object detection on large-scale LiDAR point cloud.

{\small
\bibliographystyle{ieee_fullname}
\bibliography{cvpr2021}
}

\end{document}